# Structural Alignment for Reliable Industrial AI: Bridging Physical Reality, Data, Models, and Human Intent

Lizhi Xiao[1,4§*], Sihong Wu[2§*], Victoria Xiao[3], Yiqiao Song[4]

Chen Gu[5], Jianwei Ma[6], Xinming Wu[7], Aimé Fournier[2]

## Abstract

Artificial intelligence is increasingly deployed in critical industrial domains, including healthcare, energy grids, subsurface exploration, where failures can have severe consequences for human safety, system stability, and economic outcomes. Yet AI is still evaluated primarily through benchmark accuracy, a model-centric metric that fails to capture the structural complexity and risks of real-world deployment. We propose a framework that views industrial AI reliability as a problem of structural alignment across four interacting worlds: physical, representational, machine, and human cognitive. These worlds are connected through two interfaces: digitalization, linking physical reality to computational representations, and goal encoding, translating human cognition to the machine objectives. Together, they define the space of admissible solutions. We characterize the solution space through four attributes: existence, non-uniqueness, robustness, and interpretability and show how mismatches arise at interfaces and propagate across worlds to produce reliability failures. Applications to healthcare, energy grids, and subsurface exploration illustrate that although dominant failure modes differ across domains, for example, interpretability in healthcare, robustness in energy grids, and non-uniqueness in subsurface exploration, all originate from a shared structural mechanism. By shifting the focus from model-centric evaluation to system-level alignment, this framework offers a principled foundation for assessing and governing reliability in industrial AI systems.

## Significance Statement

Artificial intelligence is rapidly transforming critical infrastructure, from operating rooms and power grids to subsurface exploration, yet ensuring its reliability under real-world conditions remains an unsolved challenge. Unlike benchmark environments, industrial settings expose AI systems to incomplete observations, evolving contexts, and human objectives that resist simplified encoding. This study introduces a structural framework that reframes reliability as a problem of alignment across four worlds, physical, representational, machine, and human cognitive, mediated by two interfaces: digitalization and goal encoding. By tracing how mismatches arise at these interfaces and propagate through the solution space characterized by existence, non-uniqueness, robustness, and interpretability, the framework reveals a shared structural mechanism underlying failures that appear domain-specific. The result is a unified foundation for diagnosing, governing, and ultimately preventing reliability failures in industrial AI, shifting the focus from model-centric evaluation to system-level structural alignment.

## 1. Introduction

Artificial intelligence (AI) systems are increasingly deployed in industrial settings where they inform decisions with direct consequences for human safety, economic outcomes, and societal welfare (1-8). Reliability is therefore a primary requirement, especially in domains such as healthcare, energy infrastructure, and engineering operations (9-11). These systems must operate under evolving conditions, incomplete observations, and imperfectly specified objectives. Despite growing deployment, maintaining reliability in practice remains a persistent and largely unresolved challenge, with failures that are not merely academic but carry immediate, severe, and sometimes irreversible consequences (12, 13).

Recent failures highlight the urgency of this problem. In healthcare, AI-enabled surgical systems have been linked to patient injuries and device recalls within months of regulatory approval (12). In energy systems, a cascading blackout affecting over 50 million people in 2025 exposed the fragility of automated control and stability mechanisms under rare but critical disturbances (13). In subsurface exploration, AI-assisted drilling continues to incur billions of dollars in losses from dry holes, where predictive accuracy masks catastrophic decision failures (11, 14). These events share a common pattern: models that perform well on standard benchmarks fail unpredictably in deployment. The pattern is not accidental — it is structural.

A growing body of work has examined the reliability, robustness, explainability, and governance of AI systems, drawing attention to fundamental challenges that are often overlooked (15-29). However, industrial AI systems are still typically assessed through model-centered performance metrics such as accuracy, prediction error or benchmark scores, favoring tractable and readily measurable proxies over broader evaluation of reliability and operational performance (26, 30-34). Persistent gaps between benchmark performance and real-world outcomes suggest that model capability alone cannot fully explain reliability (34-36). More fundamentally, AI systems are evolving from isolated predictive models into increasingly interactive socio-technical systems that link human decision-making, computational processes, data representations, and physical environments (Figure 1). As a result, reliability emerges not only from model behavior but also from the interactions among these components. Focusing on models in isolation obscures important sources of failure and limits effective governance.

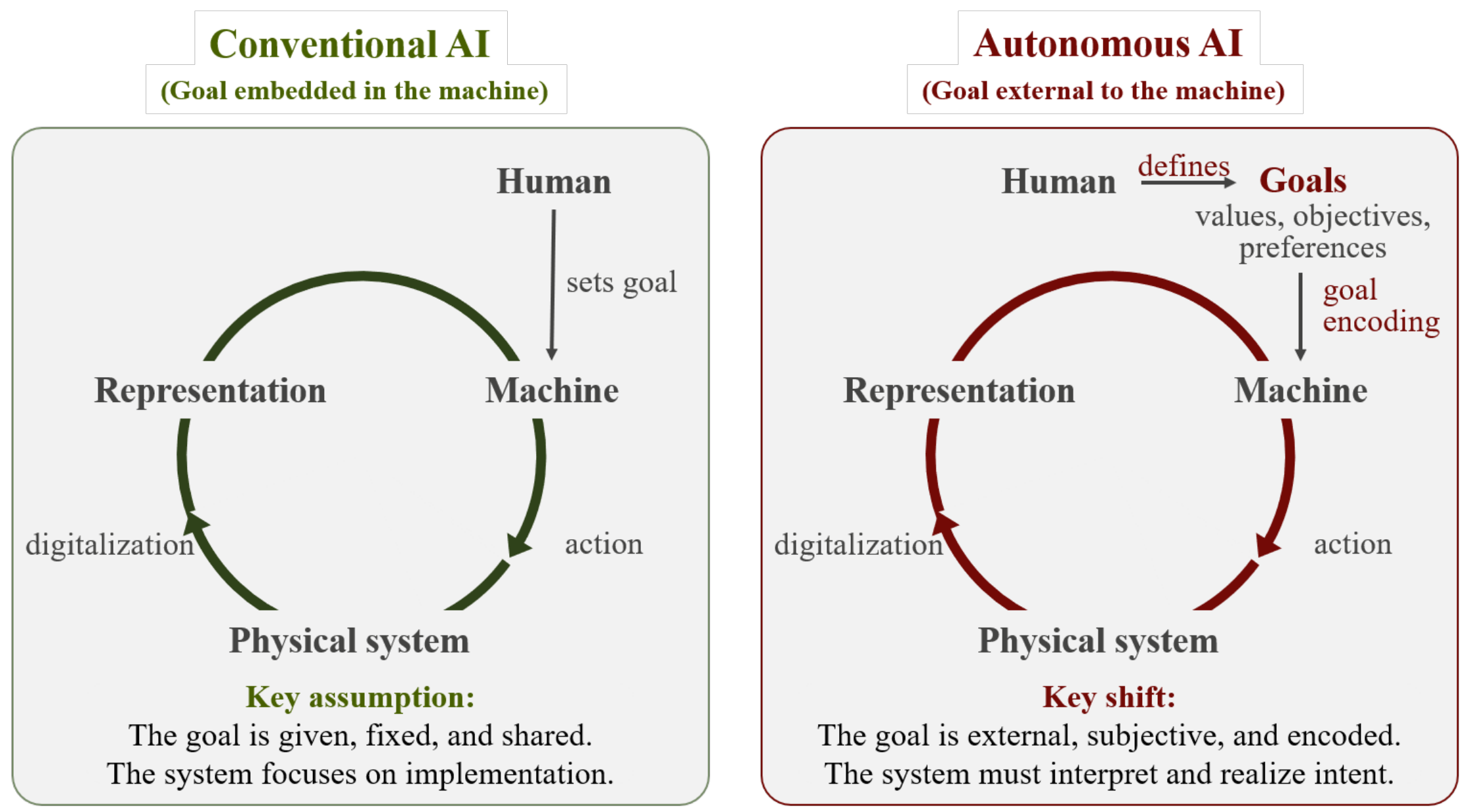

**Figure 1.** Evolution of AI systems toward greater autonomy and interaction.

This broader perspective highlights a central question: where do reliability failures originate within the system? Three interrelated factors are particularly important. First, the digitalization of physical reality, through sensing, measurement, and feature construction, inevitably loses information and introduces bias (37, 38). What is not measured cannot be learned, and what is misrepresented leads to spurious correlations. Second, the encoding of human goals into optimization objectives is necessarily incomplete, trading off multiple, often competing intentions against simplified proxies (39, 40). Third, these two interfaces interact: incomplete representations expand the space of plausible solutions, while misspecified objectives determine which solutions are selected (41, 42). The result is a solution space whose structure, whether valid solutions exist, whether multiple solutions are indistinguishable, how stable they are under perturbation, and whether they can be meaningfully interpreted — remains poorly understood and rarely examined.

A fundamental perspective is to view an AI system as a solution to a learning problem defined jointly by data and objectives (41, 43-47). This perspective naturally shifts attention from individual models to the structure of the solution space: whether valid solutions exist, whether multiple solutions satisfy the constraints, how stable they remain under perturbations, and whether they can be meaningfully interpreted (14, 33, 48). We summarize these properties through four structural attributes of AI solutions: existence, non-uniqueness, robustness, and interpretability. These attributes are not determined solely by model architecture and learning processes; they are fundamentally shaped by data and objectives that define the problem itself. Data arise from representations of physical reality and are inevitably incomplete and imperfect (33,37, 49, 50). Objectives translate human intentions into machine-optimizable forms and are often simplified or proxy-based (39-41, 51, 52). As a result, many well-known AI failures, including poor generalization, unstable behavior, and misalignment between optimized targets and operational goals, can be traced to mismatches between learned solutions and these upstream problem formulation (41, 42, 53).

To formalize this perspective, we propose a framework for industrial AI reliability organized around four interacting worlds: the physical, representational, machine, and human cognitive worlds (Figure 2). Two interfaces connect these worlds. Digitalization determines how physical reality is measured and represented, while goal encoding translates human intentions into formal objectives for optimization. Together, they shape the space of admissible solutions, characterized by four attributes: existence, non-uniqueness, robustness, and interpretability. In this framework, reliability emerges not from models alone but from alignment across worlds, interfaces and solution-space attributes.

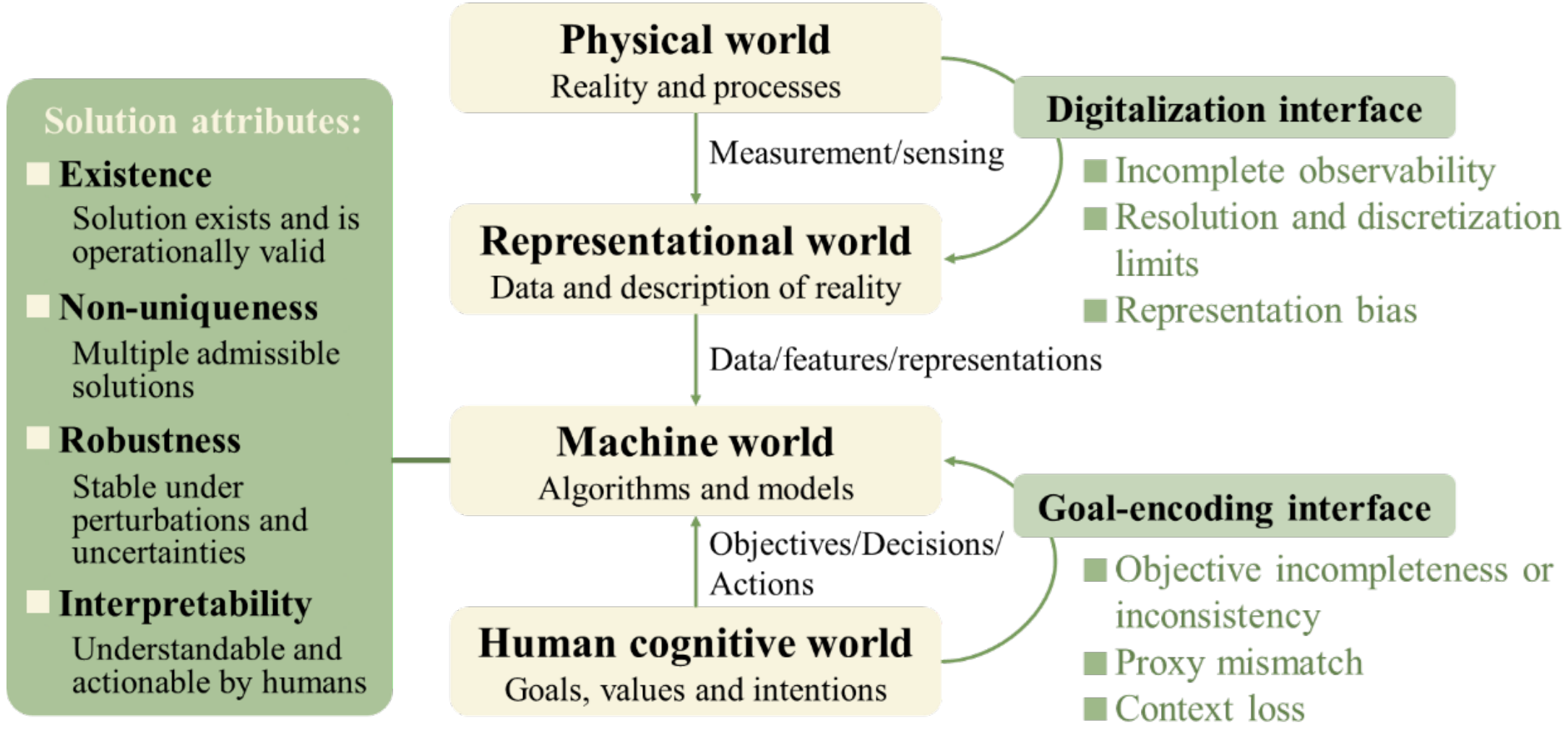

**Figure 2**. Structural framework for industrial AI reliability. The framework comprises four interacting worlds (physical, representational, machine, and human cognitive) linked by two interfaces: digitalization (mapping physical reality to representations) and goal encoding (translating human intentions into machine-optimizable objectives). These interfaces jointly shape a solution space characterized by four attributes: existence, non-uniqueness, robustness, and interpretability. Reliability is not a property

of any single model but emerges from alignment across worlds, interface, and solution-space attributes.

The framework is both diagnostic and actionable. It enables failures to be traced to their upstream origins: whether in digitalization, goal-encoding or the interaction between them. It also explains why improved model performance does not necessarily improve reliability: when the solution space is misaligned, better optimization merely selects a more polished but still structurally flawed solution. By identifying which attributes are compromised and where misalignment arises, the framework provides a basis for targeted intervention and governance.

In the sections that follow, we examine how misalignments emerge at these interfaces and shape the resulting solution space. We illustrate the framework through examples from healthcare, energy systems, and subsurface exploration, showing that although dominant failure modes differ across domains, they arise from a shared structural mechanism. This perspective shifts attention from benchmark performance to the design and governance of aligned AI systems.

## 2. A Structural Framework for Industrial AI Reliability

### 2.1 Four Interacting Worlds

Industrial AI systems operate within a broader structure connecting physical processes, data representations, computational models, and human decision-making. To capture this structure, we distinguish four interacting worlds that jointly shape the conditions under which AI systems are designed and deployed: the physical world, the representational world, the machine world, and the human cognitive world (1, 2).

The physical world consists of the real systems and processes of interest, including biological systems, engineered infrastructure, industrial operations, etc., — the reality that AI systems ultimately seek to observe, predict, or influence (10, 11).

The representational world comprises the digital abstractions of measurements, features, labels, simulations, and datasets that provide computational description of physical reality (37, 38).

The machine world includes models, algorithms, optimization procedures, and computational infrastructure through which learning and inference are performed (54, 55).

The human cognitive world encompasses domain knowledge, intentions, values,

decision processes that define the objectives and use of AI systems (40,56).

Each world imposes its own constraints and uncertainties. Physical complexity, representational limitations, computational approximations, and imperfectly specified human intentions all influence reliability. Failures often arise not from any single world in isolation, but from misalignment across them (39,41).

### 2.2 Two Critical Interfaces

***Proposition 1.*** *Industrial AI reliability depends on the upstream joint alignment of representation and goal encoding; addressing either in isolation is insufficient when misalignment exists in both.*

Across these four worlds, two interfaces play a central role in shaping the learning problem: digitalization and goal encoding.

The digitalization interface connects the physical and representational worlds. It includes sensing, measurement, discretization, and feature construction, determining which aspects of reality become observable and therefore learnable (37,51). While related efforts have emphasized data quality and measurement processes (57, 58, 59, 38), we define digitalization more broadly as the structural transformation through which continuous, high-dimensional physical processes are mapped into finite, discrete, and inevitably incomplete representations.

The goal-encoding interface connects the human cognitive and machine worlds. It translates objectives, constraints, and preferences into formal optimization targets (39,60). While related problems have been studied as objective misspecification and goal misgeneralization within specific learning paradigms (39, 52, 60), here we refer to the more general interface through which human intentions shape the selection of solutions.

Both interfaces compress complex realities into computationally tractable forms and therefore constitute primary sources of misalignment. Digitalization constrains what can be learned, whereas goal encoding determines which solutions are preferred among feasible alternatives (40,41).

### 2.3 Four Attributes of AI Solution

***Proposition 2.*** *Reliability is a property of the solution space, depending on the geometry, multiplicity, and stability of admissible solutions.*

The constraints imposed by data and objectives jointly shape the AI solution space characterized by four attributes: existence, non-uniqueness, robustness, and interpretability (41, 43).

Existence concerns whether a solution satisfying observed data and operational constraints can be realized. Empirical fit alone does not guarantee operational validity; a solution may explain the available data yet fail under deployment conditions (54). Distinguishing empirical from operational existence is therefore central to reliability.

Non-uniqueness arises when multiple solutions satisfy the available data or objectives but differ in their underlying mechanisms or deployment behavior (41, 43). Such ambiguity can result from incomplete representations, model equivalence, or causal uncertainty, allowing solutions with similar training performance to exhibit markedly different outcomes in practice (42). Non-uniqueness is therefore a fundamental and often underrecognized source of reliability risk.

Robustness characterizes the stability of solutions under perturbations in data, environments, or modeling assumptions (14, 55, 61, 62). While extensively studied in the context of distributional shift and uncertainty (55, 63), multiple near-equivalent solutions may respond differently to the same perturbation, making robustness a property of the solution space rather than of any single model (41).

Interpretability concerns the extent to which a solution can be understood, validated, and acted upon within human cognitive and operational frameworks (48, 64, 66, 67). It depends on alignment between learned representations and causal or meaningful domain structures and is widely recognized as a requirement for trust and accountability in high-stakes settings (64, 65, 68). Interpretability also functions as a constraint on the solution space by eliminating solutions incompatible with domain knowledge or operational reasoning, thereby mitigating non-uniqueness (48).

### 2.4 Relationship to Existing Frameworks

The proposed framework builds on, but departs from, several existing lines of work.

Trustworthy and responsible AI studies identify key properties such as fairness, robustness, and transparency (17, 25, 27, 61), but typically treat them as attributes of models or outputs. Here, we trace these properties to their structural origins, linking them to upstream interfaces and cross-world interactions.

Data-centric and scientific machine learning approaches emphasize data quality, measurement fidelity, and physical consistency as determinants of model behavior (26, 38, 54, 61). We extend this perspective by embedding data within the broader representational world, explicitly connecting it upstream to physical processes and downstream to human objectives.

Work on underspecification, uncertainty quantification, and distribution shift shows that similar-performing models may behave very differently under deployment conditions (41, 55). We incorporate this insight by treating non-uniqueness as an intrinsic attribute of the solution space and linking it to both digitalization and goal-encoding interfaces as its upstream sources.

In this sense, the contribution of this study is not the introduction of entirely new concepts, but their integration into a unified structural framework that connects these strands and reveals the shared mechanism through which misalignments arise across diverse AI systems and industrial contexts (28, 29).

## 3. Interfaces as Mechanisms of Reliability

### 3.1 Digitalization: From Physical Reality to Representation

***Proposition 3.*** *Digitalization defines the feasible solution space; reliability degraded by representational limitations cannot be fully recovered through downstream model optimization.*

The digitalization interface transforms continuous and often high-dimensional physical phenomena into finite representations through sensing, measurement, sampling, discretization, and feature construction (37,51). This transformation is necessarily selective. Information is lost, uncertainty is introduced, and prior assumptions become embedded in the representation itself.

Three sources of misalignment commonly arise at this interface: incomplete observability, where relevant variables remained unmeasured or unobservable; resolution and discretization limits, where important temporal or spatial details are not preserved; and representation bias, where data processing or feature construction emphasizes certain patterns while suppressing others (37, 69).

From the structural perspective, limitations in digitalization directly shape all four attributes by restricting the information available to the learning process. First, digitalization restricts existence. Poorly resolved or incomplete data may admit empirical solutions that fit the representation well, yet fail to reflect physically meaningful or causally valid behavior (54). The learning problem may appear solvable within the data representation while lacking an operationally valid solution with respect to the underlying system. Second, these limitations increase non-uniqueness.

Informational ambiguity in the mapping from physical states to data makes distinct states indistinguishable, while noise and preprocessing further blur distinctions, resulting in multiple solutions that are equally consistent with the available data (41, 43). Third, these limitations undermine robustness. Models may capture local regularities rather than relationships that persist across conditions, leading to sensitivity to distribution shift and degraded performance under changing environments (42, 55). More fundamentally, the absence of stable causal structure limits generalization, while different solutions within the non-unique set may rely on different features, resulting in inconsistent and unpredictable behavior under perturbations. Finally, digitalization also shapes interpretability. Misalignment between digital representations and human semantic structures complicates interpretation. Systematic distortions arising from biased or incomplete data can yield explanations that appear internally coherent but are fundamentally misleading (48,64). Moreover, digitalization inevitably embeds prior assumptions, creating a risk of circularity and an illusion of explanatory depth (65).

### 3.2 Goal Encoding: From Human Intent to Optimization

***Proposition 4.*** *When multiple admissible solutions exist, goal encoding governs selection among them; reliability depend on both the diversity of admissible solutions and the objectives used to distinguish among them.*

The goal-encoding interface governs how human intentions are translated into machine-optimizable objectives, through the design of loss functions, reward signals, and evaluation criteria (39,60).

A central limitation of goal encoding is that formal objectives are necessarily simplified and may introduce systematic mismatch. Three characteristic sources arise: objective incompleteness or inconsistency, where relevant constraints or trade-offs are omitted or incompatible; proxy mismatch, where surrogate metrics fail to capture the intended goal; and context loss, where deployment conditions are not reflected in the optimization problem (38,40). Such divergence between underlying intention and encoded objective has been shown to lead to systematic failures, variously described as objective misspecification or goal misgeneralization (39, 52).

From a structural perspective, goal encoding shapes all four attributes by determining how optimization is directed and how solutions are selected from the feasible set. First, goal encoding restricts existence when the specified objective is incompatible with the available data, model capacity, or task structure. The learning problem may admit no feasible solution when the specified objective cannot be jointly satisfied with the available data and task constraints (56). Second, goal encoding shapes non-uniqueness by defining how optimization proceeds and how solutions are distinguished — a role especially consequential when multiple solutions appear equivalent. Incomplete objectives or proxy metrics can flatten the loss landscape, allowing distinct solutions to achieve comparable objective values (41). By defining evaluation criteria, it determines which differences in generalization behavior, causal structure, and robustness are recognized or overlooked. Third, goal encoding shapes robustness by determining which patterns are rewarded during optimization. Objectives that prioritize accuracy over stability may lead models to exploit spurious relationships that fail under distributional shift (42). Context loss further widens the gap between training objectives and deployment conditions, degrading operational performance (55). Finally, goal encoding shapes interpretability by determining how causal or human-understandable constraints are incorporated into the objective. Proxy metrics may yield explanations that are causally uninformative. Under non-uniqueness, such objectives leave logically distinct but equivalently performing solutions indistinguishable (64).

### 3.3 Interaction Between Digitalization and Goal Encoding

Digitalization and goal encoding jointly constrain AI system behavior by defining the feasible solution space and the selection among solutions (39, 41). Reliability failures often emerge from their interaction. Widely observed phenomena such as distributional shift, shortcut learning, and objective misspecification can frequently be understood through this lens (42, 55). Two general patterns arise:

Compounded ambiguity occurs where representational limitations enlarge ambiguity within the solution space, while objective mismatch guides optimization toward unstable or misaligned solutions. In this case, uncertainty introduced by digitalization is amplified through goal encoding.

Hidden infeasibility occurs when critical real-world constraints are absent from both the representation and the objective. Solutions may appear valid within the learning problem yet fail under deployment conditions. In this case, both interfaces jointly sustain an illusion of feasibility (38, 40).

### 3.4 Diagnostic Implications and Governance

***Proposition 5.*** *Effective governance requires sustained control over both interfaces, their interaction, and lifecycle dynamics.*

Viewing digitalization and goal encoding as structural interfaces provides a basis for systematic diagnosis of reliability failures. This perspective enables analysis of where misalignment arises, which solution attributes are affected, and how failures propagate across domains. It also reveals why improvements in model performance do not necessarily improve reliability when failures originate upstream. Effective governance therefore requires interventions targeted at the source of misalignment, while failures arising from their interaction require coordinated redesign across both.

Governance of digitalization should ensure that representations preserve the causal structure, variability, and uncertainty of the underlying system (54). This requires careful design of observability, representation, and monitoring over time. Observability design should prioritize causally relevant variables and ensure coverage of the full range of system variation, including rare but operationally critical conditions. Data transformations and feature construction should preserve physically meaningful structure and avoid operations that obscure causal signals or introduce systematic bias (37, 70). More fundamentally, uncertainty should be characterized explicitly. Incomplete observability, measurement error, and representational limitations inevitably introduce uncertainty into industrial AI systems. Governance should therefore ensure that uncertainty is quantified, communicated, and propagated through downstream decision-making rather than suppressed through single-point predictions (14). Continuous monitoring helps detect representation drift as system conditions evolve (55). Throughout, the digitalization pipeline should remain transparent and traceable, with documentation of sampling protocols, transformation steps, and known sources of information loss to support auditing, reproducibility, and downstream interpretation.

Governance of goal encoding should ensure that objectives faithfully represent operational intent, remain stable over optimization, and adapt to evolving contexts (40, 52). This requires control over objective design, optimization behavior, and lifecycle alignment. Objectives should reflect not only performance targets but also safety requirements, risk tolerances, and operational constraints. They generally combine both numerical objectives that can be directly optimized and conceptual objectives that describe desired system properties. The latter often require translation into proxy metrics, constraints, or governance mechanisms, creating a major source of goal-encoding mismatch. Trade-offs should be explicitly represented when multiple goals are present (56). Proxy metrics should be used with caution and their limitations acknowledged. Normative constraints, such as safety, ethical, or regulatory requirements, should be encoded as hard constraints. Optimization processes should be monitored to prevent reward exploitation or unintended behaviors, particularly under incomplete or proxy-based objectives (39). Incorporating constraints that promote stability, invariance, or causal consistency can reduce sensitivity to distributional shifts. Objectives should be stress-tested under diverse conditions, including out-of-distribution and adversarial scenarios (55). Finally, objectives should be regularly reviewed and updated to remain aligned with evolving operational contexts. Transparency in objective specification, including documentation of assumptions, constraints, and trade-offs, is essential to support interpretability, accountability, and reliable system behavior.

More broadly, governance should focus on the interfaces through which misalignment enters the system. Guardrails on digitalization determine what information may be lost, transformed, or represented, whereas guardrails on goal encoding determine how objectives may be specified, approximated, and optimized. Together, these guardrails define the boundaries within which representations and objectives remain aligned with operational reality, thereby limiting pathways to reliability failure.

## 4. Solution Attributes and the Structure of Reliability

### 4.1 Propagation and Interactions Among Attributes

Beyond performance metrics, the proposed framework characterizes industrial AI reliability through four structurally coupled attributes: existence, non-uniqueness, robustness, and interpretability. Their interdependencies take the form of causal, trade-off, and constrained relations, which together generate the feedback loops that shape system-level reliability and explain why improving one attribute may inadvertently degrade another (Figure 3).

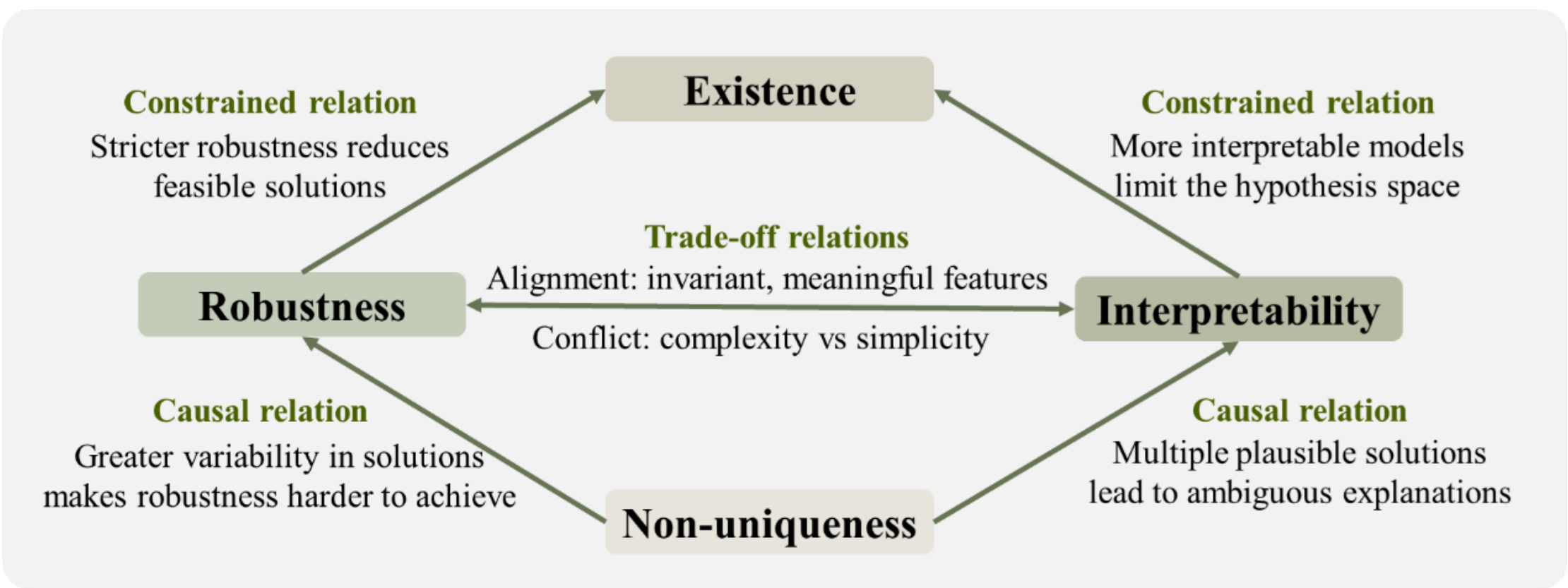

**Figure 3.** Relationships among solution attributes.

**Causal relations.** Non-uniqueness degrades both robustness and interpretability (Figure 4). When multiple solutions achieve comparable training performance, they may rely on fundamentally distinct mechanisms and thus respond differently under deployment conditions (41, 43). Information can be represented and organized in fundamentally diverse ways across solutions. As a result, robustness varies across the near-equivalent solution set, and interpretability becomes solution-dependent, with only causally consistent solutions yielding reliable explanations (64).

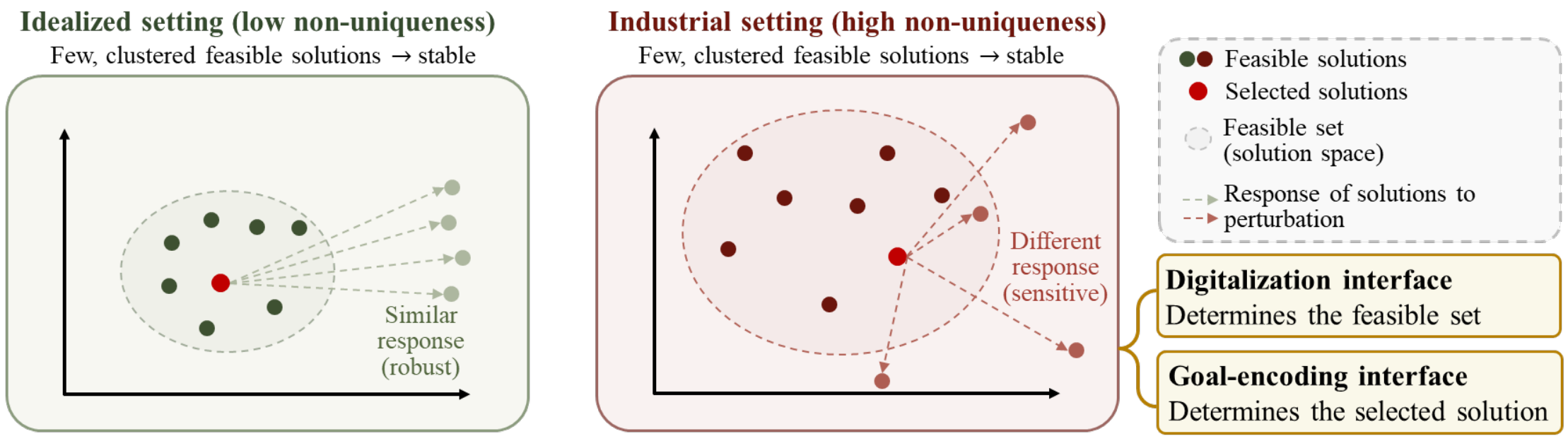

**Figure 4.** The structure of the solution space governs reliability. The solution space is bounded by digitalization (which determines representational completeness) and shaped by goal encoding (which selects among admissible solutions). Reliability fails when the solution space is underconstrained (high non-uniqueness) or when the objective selects a solution that is unstable or misaligned.

***Proposition 6.*** *Robustness and interpretability are not properties of a single solution but emergent properties of the solution space; reliability failures often arise from the coexistence of multiple admissible solutions with heterogeneous responses under perturbation.*

**Trade-off relations.** Robustness and interpretability are coupled and may either reinforce or constrain one another depending on the representations learned. Alignment occurs when models capture relationships that persist across environments and that also correspond to human-meaningful concepts — invariant, semantically coherent features that support both generalization and explanation (48, 54). However, this alignment is not always guaranteed. Robustness may rely on complex or distributed representations that lack clear semantic correspondence, while enforcing interpretability can impose structural simplifications that reduce representational capacity and limit robustness under distribution variation (55, 64). This trade-off reflects a broader tension between representational richness and structural simplicity (70, 71).

***Proposition 7.*** *Robustness and interpretability are neither uniformly aligned nor universally opposed; their relationship depends on whether invariant patterns correspond to human-meaningful structure.*

**Constrained relations.** Robustness and interpretability each impose constraints on the solution space and, in some cases, challenge existence. Enforcing stability across environments requires models to capture invariant relationships (55), while interpretability imposes structural restrictions that limit representational capacity (64). When these constraints exceed what the data, model class, or problem formulation can support, no solution may satisfy both accuracy and imposed requirements. Existence and non-uniqueness are jointly determined by the strength of such constraints. Non-uniqueness presupposes existence, as it arises only when multiple solutions satisfy the objective. Looser constraints enlarge the feasible set and increase non-uniqueness, whereas stricter constraints reduce non-uniqueness but risk eliminating feasible solutions altogether (41). Constraint strength therefore governs a trade-off between non-uniqueness and existence.

**Feedback loops.** The interactions above form reinforcing and balancing feedback loops. A reinforcing loop arises when loose constraints increase non-uniqueness, leading to greater variability in robustness. If evaluation relies solely on training performance, solutions that exploit spurious correlations (42) may still appear adequate, providing no incentive to tighten constraints and further amplifying fragility. A balancing loop operates when robustness requirements impose stronger constraints, reducing non-uniqueness and stabilizing behavior. However, overly stringent constraints may eliminate feasible solutions and challenge existence, signaling that requirements are unattainable and prompting relaxation. These loops indicate that reliability is an emergent property of the overall constraint structure rather than of any single model. Effective governance therefore requires balancing these feedback dynamics and coordinating all four attributes.

### 4.2 Governance Implications

The structural interactions among the four attributes have direct implications for governance. Because existence, non-uniqueness, robustness, and interpretability are coupled, interventions targeting one attribute inevitably affect the others. Governance must therefore be understood as shaping the solution space rather than optimizing individual attributes in isolation.

Existence governance aims to ensure that model outputs constitute valid representations of the system under deployment conditions. This requires clear admissibility criteria, defining acceptable outputs in terms of physical feasibility, operational executability, and semantic consistency with domain knowledge (54). Representation adequacy must be assured: sufficient observability and coverage of the underlying system are necessary conditions for existence, and failures arise when key variables or critical operation conditions are not captured. Domain constraints, derived from physical laws, engineering limits, operational procedures, and regulatory requirements, should be embedded in model structures, learning objectives, or downstream decision rules (40).

Validity should be verified under deployment conditions via feasibility checks, causal validation, and coverage testing across operating regimes (14). When validity cannot be established, systems should abstain or escalate to human oversight (71).

Non-uniqueness governance aims to reduce non-uniqueness and make it explicit, avoiding false certainty from entering decision processes. Non-uniqueness often arises under empirical existence, where multiple solutions satisfy the data and objectives, while only a subset may remain valid under operational constraints (41). Governance begins with identifying the sources of ambiguity across system interfaces: whether arising from incomplete observations or underspecified objectives (43). It then proceeds by structuring the solution space: incorporating causal knowledge, physical constraints, and domain regularities helps organize plausible solutions without eliminating reasonable alternatives (54). Non-uniqueness must be made explicit, represented as distributions, ranges, or alternative scenarios rather than single deterministic predictions, so that downstream decision-makers are aware of the degree of ambiguity present (14). In this context, governance requires awareness of selection mechanisms, recognizing how decisions are made among alternative solutions and how non-uniqueness interacts with objective functions and decision rules. Finally, when non-uniqueness remains substantial, safeguards are required. Systems should defer decisions, request additional observations, or escalate to human oversight, ensuring that unresolved ambiguity is not translated into high-confidence autonomous actions (28, 71).

Robustness governance aims to define the conditions under which reliability must persist and to prevent silent degradation as those conditions evolve. Governance begins by specifying the relevant variation space, including environmental changes, sensor drift, configuration shifts, and evolving usage patterns (55). This defines the scope of robustness requirements. Robustness must then be assessed across this variation space. Evaluation should extend beyond predictive accuracy to verify that outputs remain physically feasible, operationally meaningful, and consistent with system constraints under changing conditions, and that predictions rely on stable, causally meaningful relationships rather than context-specific patterns (42). Continuous monitoring during deployment is required to detect gradual degradation arising from shifting environments or system changes (14). When robustness cannot be maintained, adaptive responses — such as recalibration, restriction of automated actions, requests for additional observations, or escalation to human oversight — should be triggered to prevent uncontrolled behavior (71).

Interpretability governance aims to ensure that meaning is reliably transferred across the machine-human cognitive interface, rather than replaced by an illusion of understanding. Interpretability depends on alignment between representations, objectives, and human cognitive structures (48). Governance begins by aligning model representations with domain semantics, ensuring that variables and features correspond to meaningful system concepts (64). It requires context-appropriate explanation mechanisms: operators, engineers, and regulators require different forms of explanation, and effectiveness depends on both contextual relevance and fidelity to the underlying model behavior. Interpretation protocols should define how outputs and explanations are to be used, clarifying their meaning, scope, and limitations to prevent over-interpretation and unwarranted confidence (65). Finally, interpretability should be supported by institutional oversight, including documentation, audit trails, and review processes for high-consequence decisions.

These governance considerations are inseparable. The attributes are structurally coupled, and governance actions on one dimension propagate effects across the others. Effective governance requires treating reliability as a system-level outcome shaped by the joint configuration of interfaces, attributes, and decision processes, actively managing the feedback dynamics among attributes and maintaining coordinated oversight across the full pipeline from physical measurement to operational decision.

## 5. Industrial AI Failures

Industrial AI failures often appear domain-specific, reflecting differences in data, objectives, and operational contexts. From a structural perspective, however, many of these failures arise from common mechanisms. The cases examined below illustrate how misalignments at the digitalization and goal-encoding interfaces propagate through the solution space and ultimately manifest as reliability failures (Figure 5). Together, they demonstrate the generality of the proposed framework as a tool for diagnosis and governance.

| Domain | Role of AI | Dominant attribute | Interface failure | Failure mode | Consequence | Governance focus |
|---|---|---|---|---|---|---|
| Healthcare | Decision support in clinical workflows | Interpretability | ▪ Representation not aligned with physiology<br>▪ Objectives misaligned with multi-objective clinical intent | ▪ Misleading or unaccountable clinical decisions | ▪ Patient harm<br>▪ Erosion of clinical accountability | ▪ Align representations with domain semantics<br>▪ Ensure safety, interpretability, and accountability |
| Energy grid | Real-time monitoring, forecasting, and control of grid infrastructure | Robustness | ▪ Incomplete and delayed state representation<br>▪ Objectives prioritize efficiency over stability | ▪ Failure to detect emerging instability<br>▪ Unresponsive behavior under rare but critical disturbances<br>▪ Cascading failures | ▪ Large-scale grid disruption<br>▪ Safety and economic consequences at societal scale | ▪ Ensure observability<br>▪ Prioritize stability<br>▪ Validate under extreme conditions |
| Subsurface investigation | Inference of underground structure and properties under uncertainty | Non-uniqueness | ▪ Indirect, band-limited observations<br>▪ Objectives optimize predictive accuracy rather than decision outcomes | ▪ Underdetermined subsurface models<br>▪ High-cost and high-risk decisions | ▪ Economic loss<br>▪ Safety risk | ▪ Quantify uncertainty<br>▪ Integrate multi-physics data<br>▪ Design decision-aware objectives |

**Figure 5** Structural comparison of AI reliability failures across healthcare, energy system, and subsurface investigation.

### 5.1 Healthcare: Goal Encoding and Interpretability Constraints

AI systems in healthcare are increasingly used to predict diagnoses, risks, and treatment outcomes across clinical settings (9, 72-76). Failures carry direct consequences for patient safety, clinical accountability, and institutional trust. The framework reveals that interpretability is the dominant governance concern in this domain, yet it is fundamentally shaped by both the digitalization and goal-encoding interfaces.

At the digitalization interface, clinical data are inherently incomplete and heterogeneous, shaped by diagnostic routines, documentation practices, and measurement conventions that vary considerably across institutions, practitioners, and time (69, 77). Models may therefore learn institutional regularities or procedural artefacts rather than physiological relationships. This representational bias is often invisible to standard performance metrics. From the framework's perspective, such misalignment directly limits interpretability: explanations generated by models may reflect workflow conventions rather than clinically meaningful reasoning, making them difficult to validate and prone to misinterpretation (48, 64). It also increases non-uniqueness: multiple solutions that fit the observed data equally well may rely on different spurious correlations, leading to divergent predictions under distributional shift (e.g., when a model is deployed in a different hospital system) (41). Governance at this interface must therefore ensure alignment between representations and underlying physiological mechanisms — through explicit separation of clinical processes from physiological signals, maintenance of data provenance, cross-context consistency checks, and continuous monitoring for representational drift as clinical practices evolve (69).

At the goal-encoding interface, clinical decision-making is inherently multi-objective and context-dependent. Translating trade-offs among treatment benefit, patient safety, long-term outcomes, resource constraints, and patient preferences into explicit, optimizable objectives inevitably introduces simplification and potential misalignment (38, 40). A model that maximizes predictive accuracy on a proxy label (e.g., in-hospital mortality) may disregard equally important objectives such as quality of life or avoidance of unnecessary interventions. Governance must therefore focus on aligning optimization targets with clinically meaningful outcomes. This includes specifying the operational context, assessing whether proxy labels and performance metrics reflect intended clinical goals, and examining the trade-offs embedded in objective design (56). Moreover, transparency is required about whose priorities are encoded — patients, clinicians, institutions, and payers may not share the same objectives — and these priorities must be revisited as medical knowledge, clinical guidelines, and organizational conditions evolve.

Interpretability functions as the central governance constraint in healthcare precisely because clinical decisions must be understandable, contestable, and integrated into professional accountability structures (12, 64, 78). This reliance on interpretability places implicit requirements on all other attributes. Existence requires alignment with well-defined clinical tasks and decision contexts, rather than artefacts of data availability or institutional convenience. Non-uniqueness must be made explicit: clinicians should be presented with uncertainty and irreducible ambiguity in medical inference, not false certainty. Robustness requires validation across institutions, populations, clinical workflows, and evolving practice conditions (55).

A paradigmatic illustration is the TruDi surgical navigation system (79). Following a machine-learning update in 2021, reported failures increased sharply — from only seven incidents prior to AI integration to over 100 malfunctions and more than 10 patient injuries by 2025 (80, 81). The primary failure involved incorrect localization of surgical instruments, leading to severe complications including cerebrospinal fluid leaks, skull base punctures, and vascular injuries. Analyzed through the framework, these failures reflect compounded misalignment at both interfaces: missing anatomical context in the data representation (digitalization failure leads to increased non-uniqueness), insufficient encoding of safety objectives and an absence of communicated uncertainty (goal-encoding failure leads to false sense of existence). Together, these gaps manifested as indistinguishable high-risk states, unstable behavior under variation, absence of valid solutions in critical cases, and misleading guidance to the operating surgeon. The TruDi case is not isolated. Evidence indicates that AI-enabled medical devices exhibit higher recall rates shortly after regulatory approval, with nearly half of recalls occurring within the first year (82) — a pattern that exposes the inadequacy of existing evaluation frameworks and underscores the need for structural governance (12).

Thus, in healthcare, reliability governance must prioritize interpretability as a binding constraint, while simultaneously addressing digitalization (ensuring representations capture physiological rather than procedural structure) and goal encoding (designing multi-objective, clinically aligned targets). Without such structural alignment, even high-accuracy models will remain opaque, brittle, and potentially harmful in real-world clinical deployment.

### 5.2 Energy Grid Systems: Interaction Between Interfaces and Robustness

AI systems in energy grid operations are used for forecasting, control, and real-time dispatch of power systems (10, 13, 83). Failures in such systems can propagate rapidly through interconnected infrastructure, and the consequences of instability — including cascading failures and large-scale blackouts — are immediate, severe, and difficult to reverse.

Robustness is the dominant governance requirement in this setting, jointly shaped by both the digitalization and goal-encoding interfaces. At the digitalization interface, limitations arise from partial observability, heterogeneous measurement systems, and temporal asynchrony. These representational gaps directly constrain the solution space: states that are indistinguishable in the representation may correspond to critically different physical conditions, increasing non-uniqueness (41, 43)and undermining the system's ability to detect emerging instability. Governance at this interface must therefore ensure the adequacy and integrity of grid state representation: maintaining observability through sensor placement and redundancy, validating consistency across sensing infrastructures, safeguarding data integrity against communication failures or cyber manipulation, and ensuring that state estimation preserves the underlying network structure rather than introducing measurement artefacts (83).

At the goal-encoding interface, the challenge is equally demanding. Grid operations must balance system stability, reliability margins, economic efficiency, security constraints, and regulatory requirements. Governance must ensure that encoded objectives reflect the hierarchy of real grid priorities, with system security taking precedence over economic efficiency under contingency conditions (38, 40). This includes specifying the operational context of model use, distinguishing among tasks such as real-time control, contingency management, and economic dispatch; embedding physical and security constraints into optimization design; and aligning model outputs with operator procedures and institutional rules. Moreover, robustness requires validation across diverse regions, infrastructures, load regimes, contingencies, extreme events, and system evolution, along with mechanisms to detect performance degradation under changing sensing, demand, and network conditions (14, 55).

The Iberian blackout of April 28, 2025 provides a critical illustration of interface interaction failure at grid scale (84-86). The event resulted in a large-scale power failure affecting over 50 million people, triggered by a rapid loss of approximately 15 GW of generation capacity within five seconds and amplified through cascading failures across the interconnected grid. The event is particularly instructive because it occurred in a highly digitalized grid with extensive AI integration. Despite AI systems designed for forecasting and stability support, they failed to provide early warning during the disturbance, becoming effectively unresponsive at the moment of greatest operational need.

Analyzed through the framework, this failure reflects complementary deficiencies at both interfaces. At the digitalization interface, critical system dynamics and rare disturbance signatures — including local faults such as inverter failures propagating at millisecond timescales — were not adequately represented (13). Temporal mismatch between sensing resolution and fault dynamics, combined with loss of causal structure, reduced the system's ability to distinguish high-risk states from normal operating conditions. At the goal-encoding interface, stability and risk were insufficiently prioritized in the objective, leaving the system without adequate incentive to recognize and respond to emerging instability. Together, these interface deficiencies undermined robustness, not because the models were inherently incapable, but because the solution space was underconstrained by incomplete representations and misdirected by misspecified objectives. The failure was structural, not algorithmic.

Governance in this domain therefore requires coordinated redesign across both interfaces. On the digitalization side, observability must be extended to capture fast dynamics and rare events, and state representations must preserve causal structure. On the goal-encoding side, objectives must embed stability and risk as primary constraints, not optional trade-offs, and evaluation must be validated under extreme and out-of-distribution conditions. Without such structural governance, even highly accurate AI systems will remain brittle in the face of real-world grid disturbances.

### 5.3 Subsurface System Investigation: Intrinsic Non-Uniqueness

AI systems in subsurface investigation are used to infer underground structures and properties from non-invasive remote sensing and geophysical observations (11, 87-90). These problems are inherently ill-posed: multiple subsurface configurations can be consistent with the same surface measurements. Failures manifest as incorrect subsurface representations that propagate into high-cost or high-risk decisions — misplaced drilling, mischaracterization of reservoirs, or underestimation of subsurface hazards.

Non-uniqueness is the primary governance concern in this domain, with the digitalization interface acting as the key structural bottleneck. Subsurface states are inferred through physical operators that provide only partial, model-mediated representations, subject to fundamental limits of resolution and noise (91, 87). Incomplete observability is intrinsic and cannot be eliminated. From the framework's perspective, this representational limitation directly expands the solution space: multiple physically distinct configurations become indistinguishable in the data, increasing non-uniqueness (41, 43). Governance at the digitalization interface must therefore focus on validating acquisition quality, quantifying resolution limits, integrating multi-physics data (e.g., seismic, gravity, electromagnetic, well logs) to reduce ambiguity, and tracking the temporal validity of representations as subsurface conditions evolve (11). Rather than treating representational incompleteness as a defect to be hidden, it must be explicitly managed and communicated.

At the goal-encoding interface, the challenge is twofold. First, physical constraints must be incorporated into the objective to restrict the solution space, ensuring consistency with governing physics and known reservoir behavior (54). Second, the objective must balance model characterization accuracy, safety requirements, long-term reservoir integrity, economic performance, and regulatory requirements over extended decision horizons — where present inferences shape future operational states. Standard practice optimizes predictive accuracy, but the true objective is decision quality: discovering viable resources and avoiding costly dry holes. This proxy mismatch treats all prediction errors as equal, ignoring their vastly different economic, operational, and safety consequences (38, 39). A model that performs well statistically can still drive high-cost failures by treating a catastrophic mischaracterization as equivalent to a minor prediction error. Governance must shift from accuracy-based optimization to decision-aware objectives, incorporating cost asymmetry, risk tolerance, and uncertainty (14).

A canonical and economically consequential failure arising from intrinsic non-uniqueness is the dry hole — an exploration well drilled in a location that fails to contain commercially recoverable hydrocarbons, geothermal resources, or mineral deposits (**Error! Reference source not found.**). Each dry hole may cost millions of dollars, making this among the most economically significant failure modes in industrial AI applications. Analyzed through the framework: at the digitalization interface, seismic data are band-limited and resolution-constrained; gravity and electromagnetic data are diffusive and provide inherently smoothed responses; well logs measure only a narrow column through kilometers of rock; core samples provide point-scale information at high cost and sparse spatial coverage. Such observational incompleteness is intrinsic, but can be actively managed through integration of multi-physics data, analysis of resolution and sensitivity, and explicit quantification of model uncertainty. At the goal-encoding interface, a proxy mismatch between predictive accuracy and decision quality leads the optimizer to treat all errors symmetrically, whereas the true cost structure is highly asymmetric. Governance must therefore implement ensemble-based comparison of plausible subsurface models, explicit propagation of structural uncertainty into forecasts and risk assessments, and integration of uncertainty quantification into operational planning — not as a post-hoc reporting exercise, but as a core component of decision support (11, 14).

Reliability governance in this domain demonstrates the framework's logic: identify mismatches at the interfaces (digitalization: incomplete observability; goal encoding: proxy mismatch), diagnose the affected attributes (primarily non-uniqueness, secondarily robustness and interpretability), and design controls that target domain-specific structural vulnerabilities. While subsurface investigation differs radically from healthcare or energy grids in its physical processes and operational timelines, the structural mechanism of failure, and therefore the logic of governance, remains consistent.

## 6. Limitations

The framework advanced in this work is structural and conceptual rather than operational. Several limitations should be explicitly acknowledged.

First, the framework does not provide algorithmic methods for detecting or resolving interface mismatches. It clarifies what must be governed and why, but translating these structural principles into concrete metrics, diagnostic tests, and optimization procedures remains an open challenge. Bridging this gap between structural diagnosis and operational implementation is the most immediate direction for future work.

Second, instantiating the four worlds requires domain expertise that the framework structures but cannot substitute. Decomposing physical processes, representational choices, learning objectives, and human intentions into the framework's categories involves judgments that carry the risk of misspecification — particularly in complex, poorly understood, or rapidly evolving systems where the boundaries between worlds are themselves contested.

Third, the framework assumes a largely static decomposition of worlds and interfaces, whereas real industrial systems co-evolve over time. Representations change as sensing technologies improve; objectives shift as operational priorities and regulatory requirements evolve; and the boundaries between worlds can be redrawn by technological or organizational change. Although lifecycle governance is recognized throughout the framework, dynamic alignment remains underdeveloped and represents a significant theoretical extension.

Fourth, socio-technical and institutional factors are substantially abstracted away. Stakeholder conflicts, incentive structures, organizational inertia, and regulatory constraints are not explicitly modeled, yet they critically shape how interfaces are designed, enforced, and revised in practice. A fuller account of industrial AI reliability would need to integrate these political and organizational dynamics alongside the structural mechanisms described here.

Finally, the framework's empirical scope is currently illustrative. The case studies demonstrate explanatory power across diverse domains and failure types, but systematic evaluation and domain-specific instantiation represent a natural direction for future development. Testing the framework's predictive and diagnostic claims in controlled or quasi-experimental settings would be a valuable next step.

## 7. Conclusion

This paper reframes industrial AI reliability as a problem of structural alignment across four interacting worlds: physical, representational, machine, and human cognitive. Rather than treating reliability as an emergent property of model performance, we argue that it is determined upstream — by how reality is measured and encoded at the digitalization interface, and by how human intentions are translated into formal objectives at the goal-encoding interface. Together, these two interfaces define the space of admissible solutions, whose structure is characterized by four attributes: existence, non-uniqueness, robustness, and interpretability.

The framework has direct implications for reliability governance. Effective reliability governance must shift from model-centric evaluation to system-level design, where data generation, objective specification, and constraint integration are treated as foundational. Interface governance controls how mismatches enter the system; attribute governance ensures that system behavior remains valid, bounded, stable, and interpretable in practice. While the dominant sources of unreliability vary across domains — healthcare, energy grids, subsurface exploration — the underlying governance logic remains consistent: diagnose where misalignment originates, trace its propagation through interfaces and attributes, and design controls that target structural vulnerabilities rather than symptoms.

More broadly, this framework reframes AI evaluation and deployment as questions of structural alignment: whether assumptions are explicit, whether representations preserve causally relevant structure, and whether objectives faithfully reflect operational intent. Approaches that emphasize invariant structure, causal reasoning, physics-based constraints, and uncertainty quantification are not optional add-ons but central to anchoring learning systems to reality and maintaining consistency across contexts.

Three limitations, acknowledged in Section 6, point directly to future work. First, the framework remains conceptual; translating its structural principles into operational metrics, diagnostic tests, and algorithmic procedures is an open challenge. Second, the decomposition into four worlds assumes a static ontology, whereas real industrial systems co-evolve — sensing, objectives, and even the boundaries between worlds change over time. Third, institutional and socio-political factors (incentives, regulatory regimes, organizational inertia) are abstracted away but critically shape how interfaces are designed and enforced. Addressing these limitations will require not only technical advances but also cross-disciplinary collaboration between AI researchers, domain engineers, and governance scholars.

As AI systems become increasingly embedded in critical industrial settings, where failures carry immediate and severe consequences for human safety, system stability, and economic welfare, this framework provides a principled foundation for diagnosing, governing, and ultimately preventing reliability failures. It opens a new agenda for trustworthy industrial AI research and practice, one centered on structural alignment rather than benchmark chasing.

## Author Affiliations:

Lizhi Xiao[1,3,5§*], Sihong Wu[2§*], Victoria Xiao[3], Yiqiao Song[4], Chen Gu[5], Jianwei Ma[6], Xinming Wu[7], Aimé Fournier[2]

1. State Key Lab. on Petroleum Resources and Engineering, Institute of Data Science, China University of Petroleum, Beijing 102249, China

2. Department of Earth, Atmospheric, and Planetary Sciences, Massachusetts Institute of Technology, Cambridge, MA 02139, USA

3. St. Cross College, Oxford University, UK

4. School of Engineering and Applied Sciences, Harvard University, Cambridge, MA 02138, USA

5. Tsinghua University, Beijing 100084, China

6. School of Earth and Space Sciences, Institute of Artificial Intelligence, Peking University, Beijing 100871, China

7. School of Earth and Space Sciences, University of Science and Technology of China, Hefei 230026, China

**§ Indicates equal contributions**

*** Correspondence authors: xiaolizhi@cup.edu.cn; sihongwu@mit.edu**